\documentclass{article}
\usepackage{float}

\usepackage{microtype}
\usepackage{graphicx}
\usepackage{subfigure}
\usepackage{booktabs} 

\usepackage{url}
\usepackage{hyperref}

\usepackage[accepted]{icml2025}

\usepackage{amsmath}
\usepackage{amssymb}
\usepackage{mathtools}
\usepackage{amsthm}
\newcommand{\uncheckmark}{\ding{55}} 
\usepackage{pifont}

\usepackage[capitalize,noabbrev]{cleveref}

\theoremstyle{plain}

\theoremstyle{definition}

\theoremstyle{remark}

\usepackage[textsize=tiny]{todonotes}

\icmltitlerunning{TriGuard: Testing Model Safety with Attribution Entropy, Verification, and Drift}

\begin{document}

\twocolumn[
\icmltitle{TriGuard: Testing Model Safety with Attribution Entropy, Verification, and Drift}




\begin{icmlauthorlist}
\icmlauthor{Dipesh Tharu Mahato}{1}
\icmlauthor{Rohan Poudel}{2}
\icmlauthor{Pramod Dhungana}{3}
\end{icmlauthorlist}

\icmlaffiliation{1}{Center for Data Science, New York University, New York, USA}
\icmlaffiliation{2}{University of New Haven, Connecticut, USA}
\icmlaffiliation{3}{Queens College, City University of New York, USA}

\icmlcorrespondingauthor{Dipesh Tharu Mahato}{dm6259@nyu.edu}

\icmlkeywords{Machine Learning, Robustness, Interpretability, ICML}

\vskip 0.3in
]



\printAffiliationsAndNotice{}  

\begin{abstract}
    Deep neural networks often achieve high accuracy, but ensuring their reliability under adversarial and distributional shifts remains a pressing challenge. We propose \textbf{TriGuard}\footnote{Project code: \url{https://github.com/dipeshbabu/TriGuard}}, a unified safety evaluation framework that combines (1) formal robustness verification, (2) attribution entropy to quantify saliency concentration, and (3) a novel \textbf{Attribution Drift Score} measuring explanation stability. TriGuard reveals critical mismatches between model accuracy and interpretability: verified models can still exhibit unstable reasoning, and attribution-based signals provide complementary safety insights beyond adversarial accuracy. Extensive experiments across three datasets and five architectures show how TriGuard uncovers subtle fragilities in neural reasoning. We further demonstrate that entropy-regularized training reduces explanation drift without sacrificing performance. TriGuard advances the frontier in robust, interpretable model evaluation.
\end{abstract}

\section{Introduction}

Deep learning has driven transformative advances in computer vision, enabling convolutional neural networks (CNNs) and their variants to achieve state-of-the-art performance across image classification, detection, and segmentation tasks. Yet despite their predictive power, modern neural models remain brittle in the face of adversarial examples \citep{goodfellow2015explaining}, distributional shifts \citep{hendrycks2021many}, and perturbation-based attacks \citep{croce2020reliable}. These vulnerabilities pose significant concerns in safety-critical applications such as autonomous driving, medical imaging, and financial screening, where model decisions must be both accurate and trustworthy.

The central problem is that accuracy alone is not a sufficient metric for reliable deployment. A model may achieve high test accuracy yet fail dramatically when exposed to slight perturbations, either crafted adversarially or arising from natural variation. This has led to a growing emphasis on robustness evaluation and explainability as core components of model assessment \citep{hooker2020benchmark, xu2020reliable}.

\paragraph{Robustness and Attribution as Safety Axes.}

Two main axes have emerged as vital to robust and interpretable models: (1) formal verification of robustness, which guarantees prediction stability under small input changes; and (2) attribution-based interpretability, which assesses the saliency and stability of model explanations. Each axes offers partial insight: formal verification answers “Will the prediction change under bounded perturbations?” while attribution asks “What parts of the input most influenced this decision?” However, these have rarely been integrated into a unified diagnostic framework.

Recent work has attempted to measure robustness using tools like DeepPoly \citep{singh2019abstract}, CROWN-IBP \citep{zhang2022towards}, and ERAN \citep{katz2021eran}, offering formal guarantees but often limited to small networks or idealized settings. Parallelly, attribution methods such as Integrated Gradients \citep{sundararajan2017axiomatic}, SmoothGrad \citep{smilkov2017smoothgrad}, and saliency maps \citep{simonyan2014deep} have become popular in model interpretation. However, these methods suffer from instability: attributions may change drastically under imperceptible input shifts or minor weight changes, undermining trust in their explanations \citep{kindermans2019reliability}.

Even more concerning, attribution instability has been shown to correlate with model susceptibility to adversarial examples \citep{ghorbani2019interpretation}. Recent efforts like ROAR \citep{hooker2020benchmark} and AOPC \citep{samek2017evaluating} suggest that measuring attribution reliability 
not just saliency is crucial. Yet, a comprehensive framework that triangulates prediction correctness, robustness guarantees, and attribution stability remains missing.

\paragraph{Our Contribution: TriGuard.}

To address this gap, we propose TriGuard, a scalable evaluation framework that triangulates three complementary safety axes:

1. \textbf{Formal Verification}: We use adversarially bounded checks (PGD+interval bound propagation) to assess whether predictions remain invariant within an $\epsilon$-ball in input space.

2. \textbf{Attribution Entropy}: We quantify the focus of explanations by measuring entropy over normalized Integrated Gradients. Lower entropy indicates sparse, localized attention hypothesized to reflect more interpretable reasoning.

3. \textbf{Contrastive Attribution Drift}: We introduce a new metric Attribution Drift Score which captures how model explanations change between two neighboring baseline inputs. Large drifts may signal sensitivity to spurious features or reliance on unstable gradients.

\section{Methods}

In this section, we describe the components of the TriGuard framework for evaluating neural network safety. TriGuard is designed around three diagnostic axes: formal robustness verification, attribution-based entropy analysis, and contrastive attribution drift. We also describe the experimental protocols used to implement and validate each axes across multiple datasets and model architectures.

\subsection{Model and Dataset Setup}
\label{model_dataset}

Let \( f_\theta : \mathbb{R}^d \rightarrow \mathbb{R}^k \) be a deep neural network classifier parameterized by weights \( \theta \), where \( d \) is the input dimension and \( k \) is the number of output classes. Given an input \( x \in \mathbb{R}^d \), the predicted class is:

\[
\hat{y} = \arg\max_j f_\theta(x)_j
\]

We evaluate this model across standard vision datasets:

\begin{itemize}
    \item \textbf{MNIST} (gray-scale digits),
    \item \textbf{FashionMNIST} (gray-scale clothing images),
    \item \textbf{CIFAR-10} (RGB natural images).
\end{itemize}

Each model is trained for 5 epochs using Adam, and tested on both clean inputs and adversarially perturbed examples.

\subsection{Axis 1: Formal Verification of Robustness}

To evaluate whether model predictions remain stable under small perturbations, we define an \( \epsilon \)-bounded region around an input \( x \) as:

\[
B_{\epsilon}(x) = \left\{ x' \in \mathbb{R}^d : \|x' - x\|_{\infty} \leq \epsilon \right\}
\]

A model is said to be \textit{formally robust} at \( x \) if:

\[
\forall x' \in B_{\epsilon}(x): \arg\max f_\theta(x') = \hat{y}
\]

We implement a bounded adversarial check using Projected Gradient Descent (PGD) to search for a counterexample within \( B_{\epsilon}(x) \). If no such \( x' \) is found after \( T \) steps of perturbation with step size \( \alpha \), we consider the model to be \textit{locally robust}:

\begin{verbatim}
x_adv = x + alpha * sign(grad)
x_adv = clip(x_adv, x - eps, x + eps)
\end{verbatim}

where \texttt{sign(grad)} is the sign of the gradient of the loss with respect to the input.

We also use a simplified Interval Bound Propagation (IBP) style check to ensure robustness holds across all \( \epsilon \)-perturbed inputs using forward bounds, implemented as:

\[
x_{\min} = x - \epsilon, \quad x_{\max} = x + \epsilon
\]

The model passes the check if the predicted class dominates across all bounds.

\subsection{Axis 2: Attribution Entropy}

Given a model and input \( x \), we compute \textit{Integrated Gradients} (IG) as the attribution method. IG for the \( i \)-th input feature is defined as:

\[
\text{IG}_i(x) = (x_i - x_i') \cdot \int_{\alpha=0}^1 \frac{\partial f_\theta(x' + \alpha(x - x'))}{\partial x_i} \, d\alpha
\]

where \( x' \) is a baseline input (typically a zero vector or an average image). To compute this in practice, we use a Riemann approximation with \( m \) steps:

\[
\text{IG}_i(x) \approx (x_i - x_i') \cdot \frac{1}{m} \sum_{j=1}^{m} \frac{\partial f_\theta\left(x' + \frac{j}{m}(x - x')\right)}{\partial x_i}
\]

We normalize the attributions to obtain a probability distribution \( p \in \Delta^d \), and compute its entropy:

\[
H(p) = - \sum_{i=1}^d p_i \log(p_i + \delta)
\]

where

\[
p_i = \frac{|\text{IG}_i|}{\sum_j |\text{IG}_j|}
\]

and \( \delta \) is a small constant added to avoid numerical instability from \( \log(0) \).

Low entropy \( H(p) \) implies that the attribution is concentrated (potentially more interpretable), whereas high entropy suggests that the attribution is diffuse or noisy.

\subsection{Axis 3: Contrastive Attribution Drift}

We define the \textit{Attribution Drift Score} (ADS) to quantify the shift in saliency maps when comparing two semantically equivalent inputs. For a fixed test input \( x \), we compute its attribution vectors \( a^{(1)}, a^{(2)} \) using Integrated Gradients (IG) under two different baselines \( x^{(1)}, x^{(2)} \), and define:

\[
\text{ADS}(x^{(1)}, x^{(2)}) = \| a^{(1)} - a^{(2)} \|_2
\]

This score reflects how sensitive a model’s explanation is to the choice of attribution reference — serving as a proxy for explanation stability under interpretability perturbations.

In our experiments, we use two commonly accepted and semantically meaningful baselines:
\begin{itemize}
    \item A \textbf{zero baseline} (all pixels set to zero), which represents complete absence of input signal.
    \item A \textbf{blurred baseline}, approximated as a smoothed version of the original image, retaining coarse structure but eliminating fine-grained details.
\end{itemize}

We chose these two baselines due to their contrasting interpretability assumptions: zero encodes null input, while blur encodes contextual prior. This contrast highlights the model’s attribution stability under plausible explanation choices. The resulting ADS is reported for each model and dataset.

\paragraph{Baseline Robustness.}
While the two baselines above are standard, we also explore ADS using \textit{random noise} and \textit{uniform} baselines in Appendix~\ref{appendix:baseline_sensitivity}. We observe that TriGuard’s drift patterns remain consistent, validating the robustness of our metric across attribution setups. Higher ADS indicates unstable or brittle explanation logic, which may not be reflected in clean or adversarial accuracy alone.

\subsection{Entropy-Regularized Attribution Training.}
To further improve attribution stability and adversarial resilience, we propose a lightweight enhancement: \textbf{entropy regularization on input gradients} during training.

Given the input gradient \( \nabla_x \mathcal{L}(f(x), y) \), we compute a normalized vector:

\[
p = \frac{|\nabla_x \mathcal{L}|}{\sum |\nabla_x \mathcal{L}|}
\]

and apply the entropy penalty:

\[
\mathcal{L}_{\text{entropy}} = - \sum_i p_i \log(p_i + \epsilon)
\]

This encourages sparser and more localized attribution maps, which as our results show correlate with lower drift and higher verification success. The full loss becomes:

\[
\mathcal{L}_{\text{total}} = \mathcal{L}_{\text{CE}} + \lambda \cdot \mathcal{L}_{\text{entropy}}
\]

where \( \lambda \) is a small regularization coefficient (e.g., \( 0.01 \) to \( 0.1 \)).

\subsection{Additional Baselines: SmoothGrad$^2$ and CROWN-IBP.}

To strengthen empirical benchmarking, we incorporate two additional baselines into the TriGuard evaluation pipeline: \textbf{SmoothGrad$^2$} \cite{smilkov2017smoothgrad} for attribution stability, and \textbf{CROWN-IBP} \cite{zhang2021towards} for certified robustness.

\textbf{SmoothGrad$^2$} enhances attribution reliability by averaging squared saliency maps over $N$ noisy input samples. Given an input $x$, Gaussian noise $\mathcal{N}(0, \sigma^2)$ is added to produce perturbed inputs $x_i$, from which gradients $s_i$ are computed:
\[
SG(x) = \frac{1}{N} \sum_{i=1}^N s_i(x_i)^2
\]
We compute entropy and drift metrics on these aggregated maps to compare attribution stability with TriGuard's Integrated Gradients-based approach.

\textbf{CROWN-IBP} integrates convex relaxation (CROWN) and interval bound propagation (IBP) to certify output stability under norm-bounded adversarial perturbations. This method offers tighter and more scalable robustness guarantees compared to vanilla random-sampling checks. We adapt CROWN-IBP into our formal verification component to benchmark the certified safety of evaluated models.

Together, these baselines provide orthogonal perspectives on model robustness - one targeting explanation consistency, the other providing provable guarantees thereby reinforcing TriGuard's comprehensive safety assessment framework.

\subsection{Experimental Evaluation Protocol}

We evaluate \textbf{TriGuard} across 18 model-dataset combinations:

\begin{itemize}
    \item \textbf{Models}: SimpleCNN, ResNet50, ResNet101, MobileNetV3 Large, DenseNet121
    \item \textbf{Datasets}: MNIST, FashionMNIST, CIFAR-10
\end{itemize}

For each setting, we record the following metrics:
\begin{itemize}
    \item Clean accuracy on test data,
    \item Accuracy under PGD attack (\( \epsilon = 0.1 \) for MNIST \& FashionMNIST, \( \epsilon = 0.3 \) for CIFAR-10),
    \item Formal verification status (pass/fail),
    \item Attribution entropy,
    \item Attribution drift score (L2 norm)
\end{itemize}

\section{Results}

\begin{table*}[t]
\centering
\caption{
\textbf{TriGuard results with entropy regularization} across five models and three datasets. We report clean accuracy, adversarial error, attribution entropy, attribution drift score, SmoothGrad$^2$, formal verification status, and CROWN-IBP success. Entropy-regularized models tend to exhibit lower drift and better attribution stability.
}
\vspace{1mm}
\label{tab:results_with_entropy}
\scalebox{0.85}{
\begin{tabular}{lccccccc}
\toprule
\textbf{Model} & \textbf{Accuracy} & \textbf{Adv Error} & \textbf{Entropy} & \textbf{Drift} & \textbf{SmoothGrad$^2$} & \textbf{FormalVerif} & \textbf{CROWN-IBP} \\
\midrule
\multicolumn{8}{c}{\textbf{MNIST}} \\
\midrule
SimpleCNN         & 98.51\% & 1.40\%  & 5.1300 & 2.48   & 4.7327 & \checkmark   & \checkmark \\
MobileNetV3 Large & 96.93\% & 28.20\% & 4.2087 & 1.62   & 5.9797 & \checkmark   & \uncheckmark \\
ResNet50          & 98.45\% & 5.00\%  & 4.1958 & 2.74   & 5.8747 & \checkmark   & \uncheckmark \\
ResNet101         & 89.87\% & 0.20\%  & 4.1874 & 2.85   & 5.9651 & \checkmark   & \uncheckmark \\
DenseNet121       & 99.18\% & 0.00\%  & 4.3683 & 3.53   & 5.7724 & \checkmark   & \uncheckmark \\
\midrule
\multicolumn{8}{c}{\textbf{FashionMNIST}} \\
\midrule
SimpleCNN         & 90.63\% & 0.40\%  & 6.0643 & 3.43   & 5.6472 & \checkmark   & \uncheckmark \\
MobileNetV3 Large & 84.00\% & 7.60\%  & 5.1612 & 10.95  & 5.5523 & \checkmark   & \uncheckmark \\
ResNet50          & 86.61\% & 3.20\%  & 6.1503 & 2.24   & 6.2143 & \checkmark   & \uncheckmark \\
ResNet101         & 81.24\% & 1.50\%  & 5.0087 & 3.58   & 5.9525 & \uncheckmark & \uncheckmark \\
DenseNet121       & 92.45\% & 0.00\%  & 6.0763 & 6.90   & 5.5949 & \checkmark   & \uncheckmark \\
\midrule
\multicolumn{8}{c}{\textbf{CIFAR-10}} \\
\midrule
SimpleCNN         & 71.94\% & 0.40\%  & 7.4011 & 1.96   & 7.5307 & \uncheckmark & \uncheckmark \\
MobileNetV3 Large & 56.27\% & 0.10\%  & 7.3683 & 2.30   & 7.3973 & \uncheckmark & \uncheckmark \\
ResNet50          & 57.40\% & 0.00\%  & 7.3640 & 0.99   & 7.0822 & \checkmark   & \uncheckmark \\
ResNet101         & 58.66\% & 0.40\%  & 7.3504 & 0.42   & 7.3063 & \checkmark   & \uncheckmark \\
DenseNet121       & 76.21\% & 0.00\%  & 7.2730 & 4.13   & 7.4167 & \uncheckmark & \uncheckmark \\
\bottomrule
\end{tabular}
}
\end{table*}

We evaluate \textbf{TriGuard} across combinations of datasets and architectures described in Section~\ref{model_dataset}. For each configuration, we report the metrics listed in Table~\ref{tab:results_with_entropy}: clean accuracy, adversarial error, attribution entropy, attribution drift score, SmoothGrad$^2$, formal verification pass, and CROWN-IBP status. All models are trained with entropy regularization unless otherwise noted.

As shown in \textbf{Table~\ref{tab:results_with_entropy}}, models like SimpleCNN on MNIST achieve high clean accuracy (98.51\%) and also pass both formal verification checks, with notably low attribution drift (2.48) and moderate entropy (5.13). On the other hand, deeper architectures like DenseNet121 achieve high accuracy (99.18\%) but fail CROWN-IBP and exhibit higher drift (3.53), highlighting latent instability not visible through accuracy alone.

\begin{figure}[t]
\centering
\includegraphics[width=0.48\textwidth]{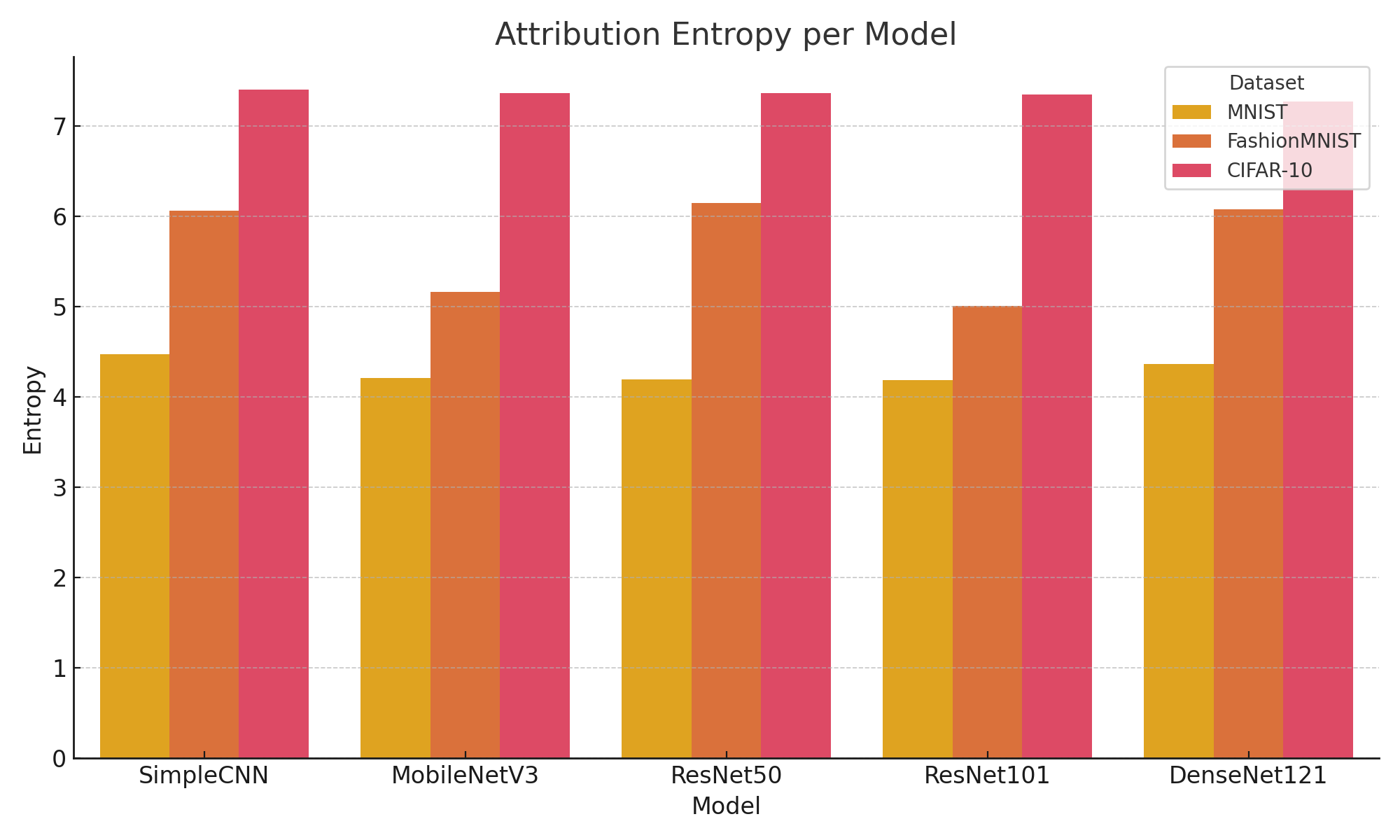}
\caption{
\textbf{Attribution entropy across all model–dataset combinations.}
CIFAR-10 models exhibit higher entropy on average, suggesting more diffuse and uncertain explanations due to input variability.
}
\label{fig:entropy_plot}
\end{figure}

\begin{figure}[t]
\centering
\includegraphics[width=0.48\textwidth]{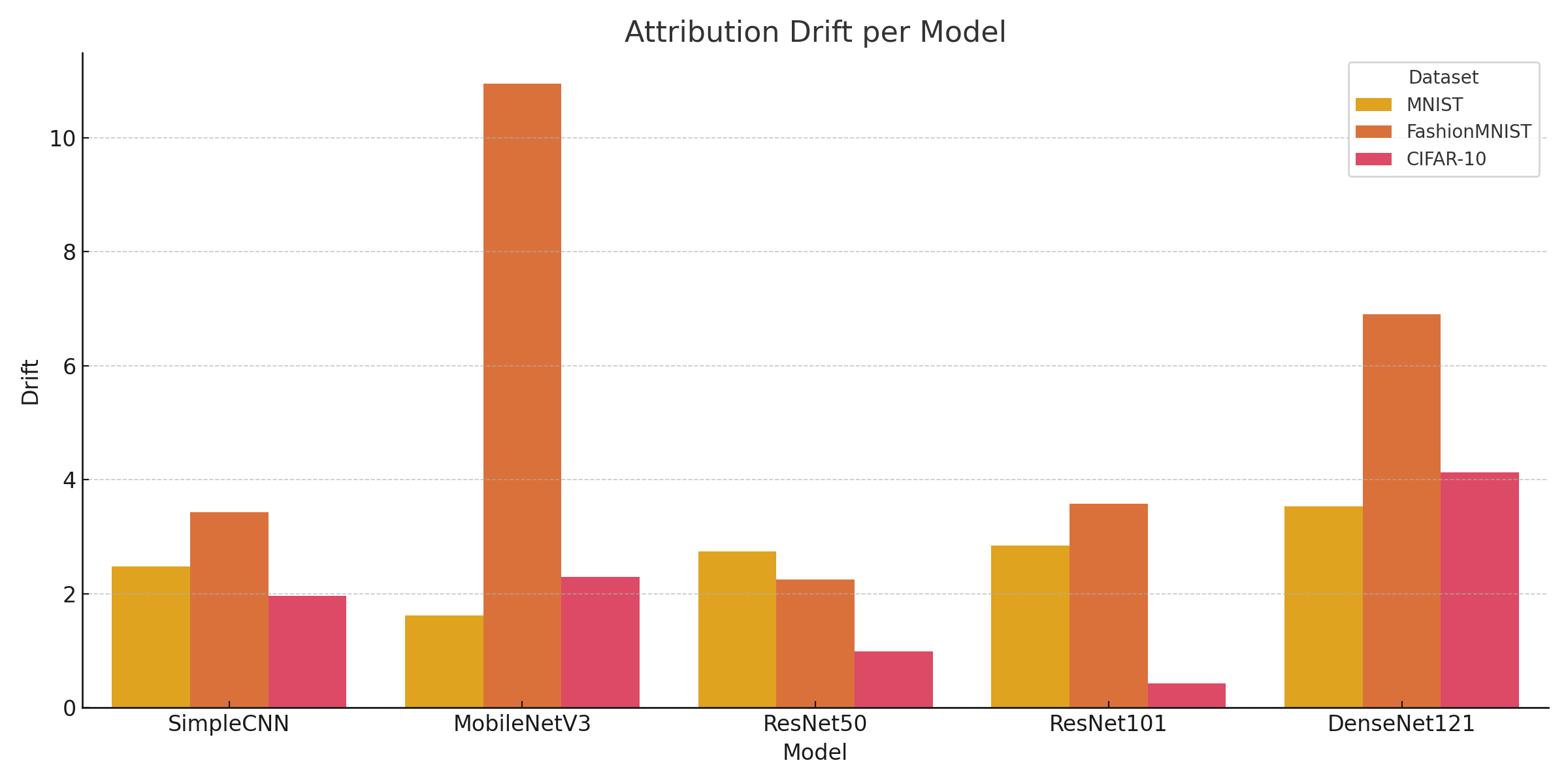}
\caption{
\textbf{Attribution drift across all models.}
FashionMNIST models show the highest drift variance, while MNIST models exhibit more stable saliency patterns.
}
\label{fig:drift_plot}
\end{figure}

\textbf{Figure~\ref{fig:entropy_plot}} presents attribution entropy across all models and datasets. We observe that CIFAR-10 models consistently yield higher entropy, indicating more diffuse attributions, likely due to input complexity. Meanwhile, \textbf{Figure~\ref{fig:drift_plot}} visualizes attribution drift, showing greater variance for FashionMNIST compared to MNIST.

\begin{figure*}[t]
\centering
\includegraphics[width=0.75\textwidth]{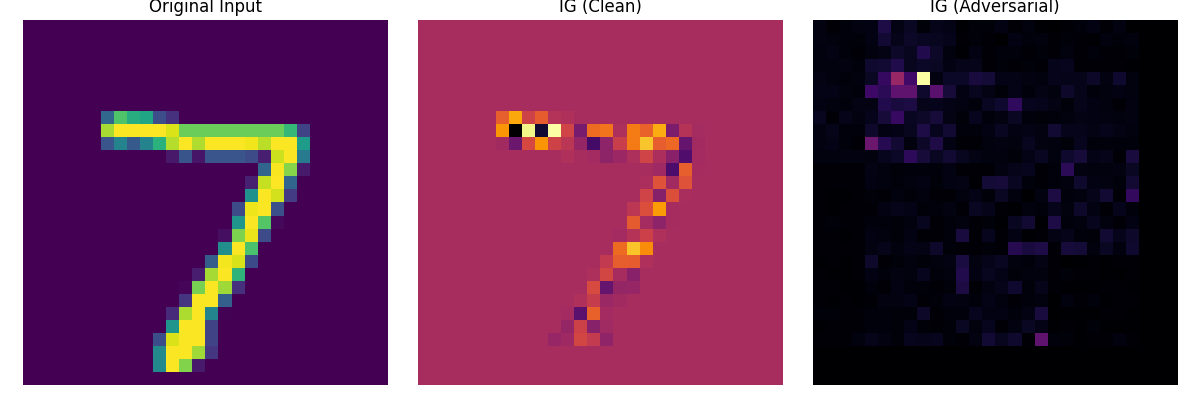}
\caption{
\textbf{Integrated Gradients (IG) saliency on clean vs. adversarial inputs.}
Visual comparison for digit “7” shows that adversarial inputs yield noisy, displaced saliency — highlighting the need for stability-aware attribution metrics.
}
\label{fig:ig_clean_adv}
\end{figure*}

\begin{figure*}[t]
\centering
\includegraphics[width=0.9\textwidth]{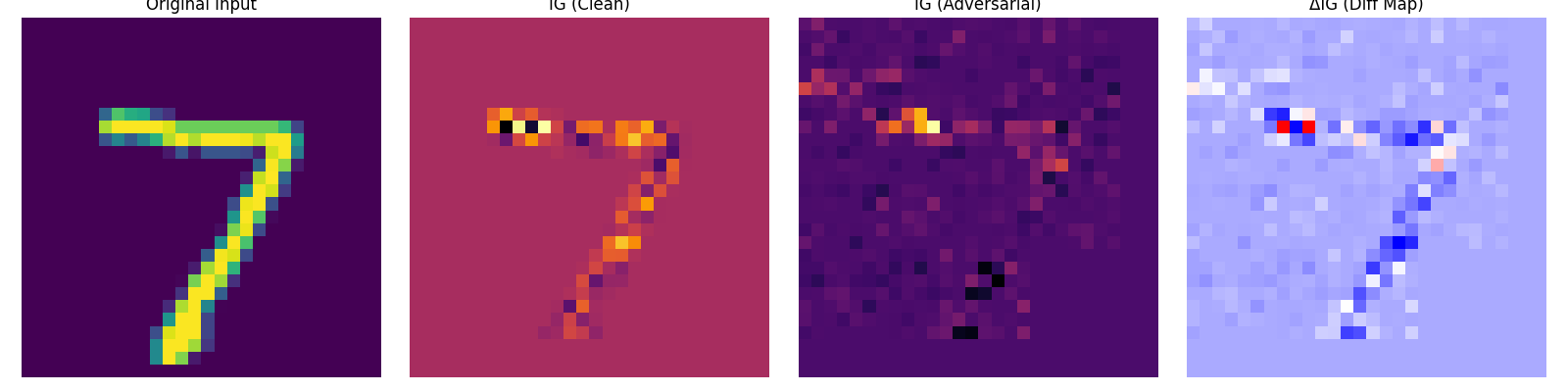}
\caption{
\textbf{Contrastive attribution map using IG under adversarial perturbation.}
Red and blue regions visualize attribution displacement, capturing explanation instability even when the prediction remains correct.
}
\label{fig:ig_delta_map}
\end{figure*}

To qualitatively compare attribution robustness, \textbf{Figure~\ref{fig:ig_clean_adv}} illustrates IG heatmaps for clean and adversarial examples of digit “7.” Adversarial perturbations produce shifted, noisy saliency, confirming fragility in explanation. \textbf{Figure~\ref{fig:ig_delta_map}} further highlights this with red/blue overlays showing attribution displacement under attack.

\textbf{Results without regularization} are shown in Appendix~\ref{appendix:unregularized_results}, reinforcing that entropy-regularized training yields more stable saliency without sacrificing accuracy.

\subsection{Correlation Analysis}

\begin{figure*}[t]
\centering
\includegraphics[width=0.31\textwidth]{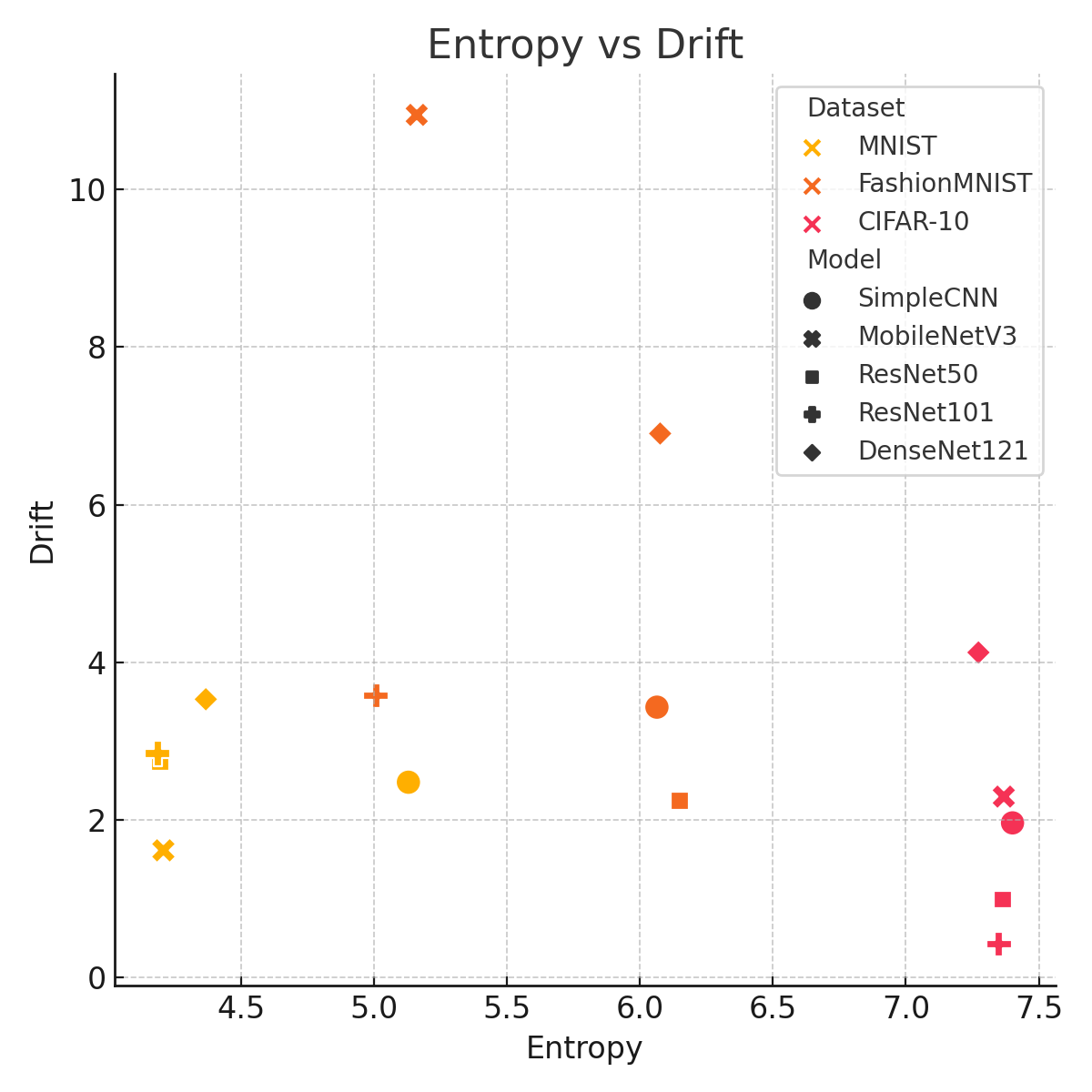}
\includegraphics[width=0.31\textwidth]{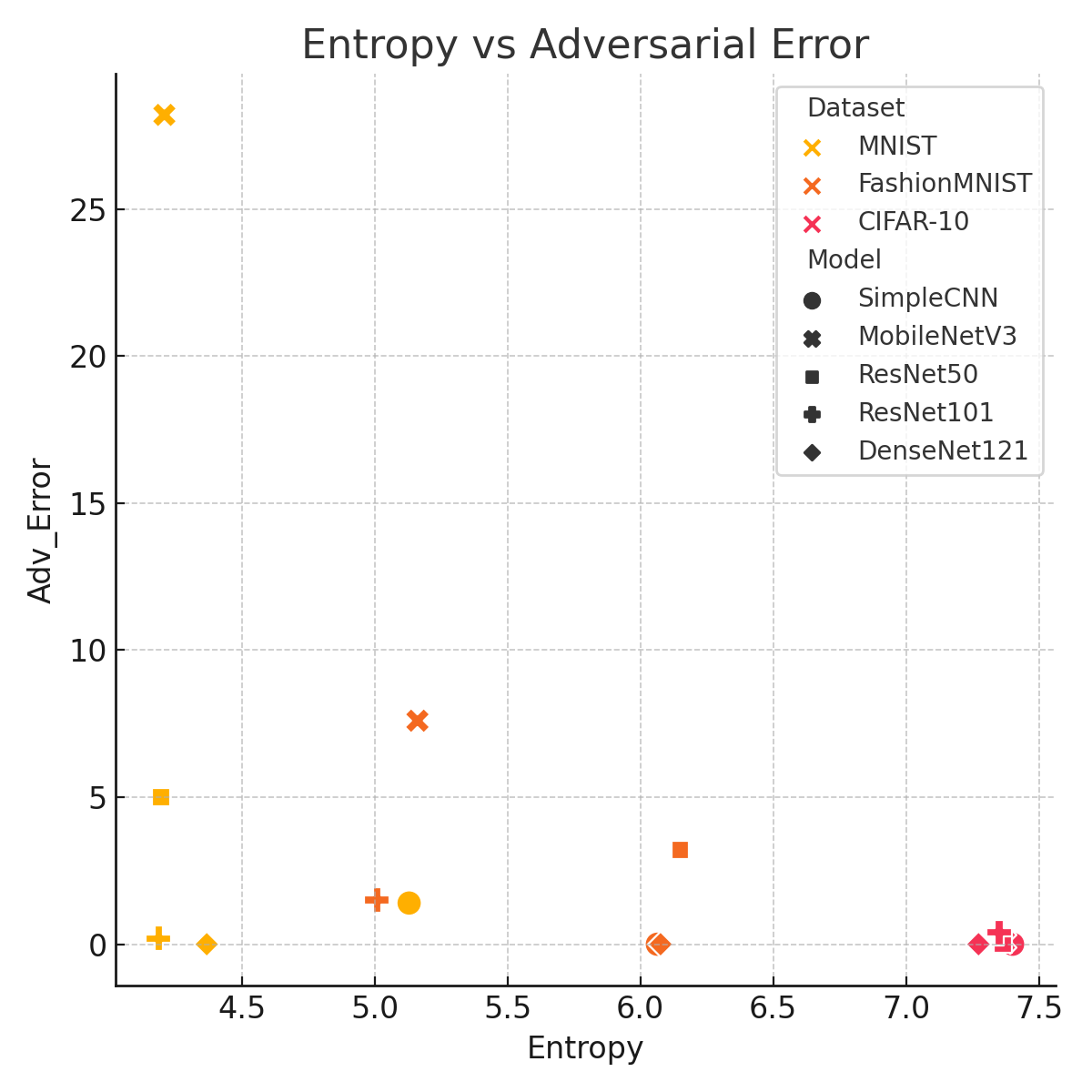}
\includegraphics[width=0.31\textwidth]{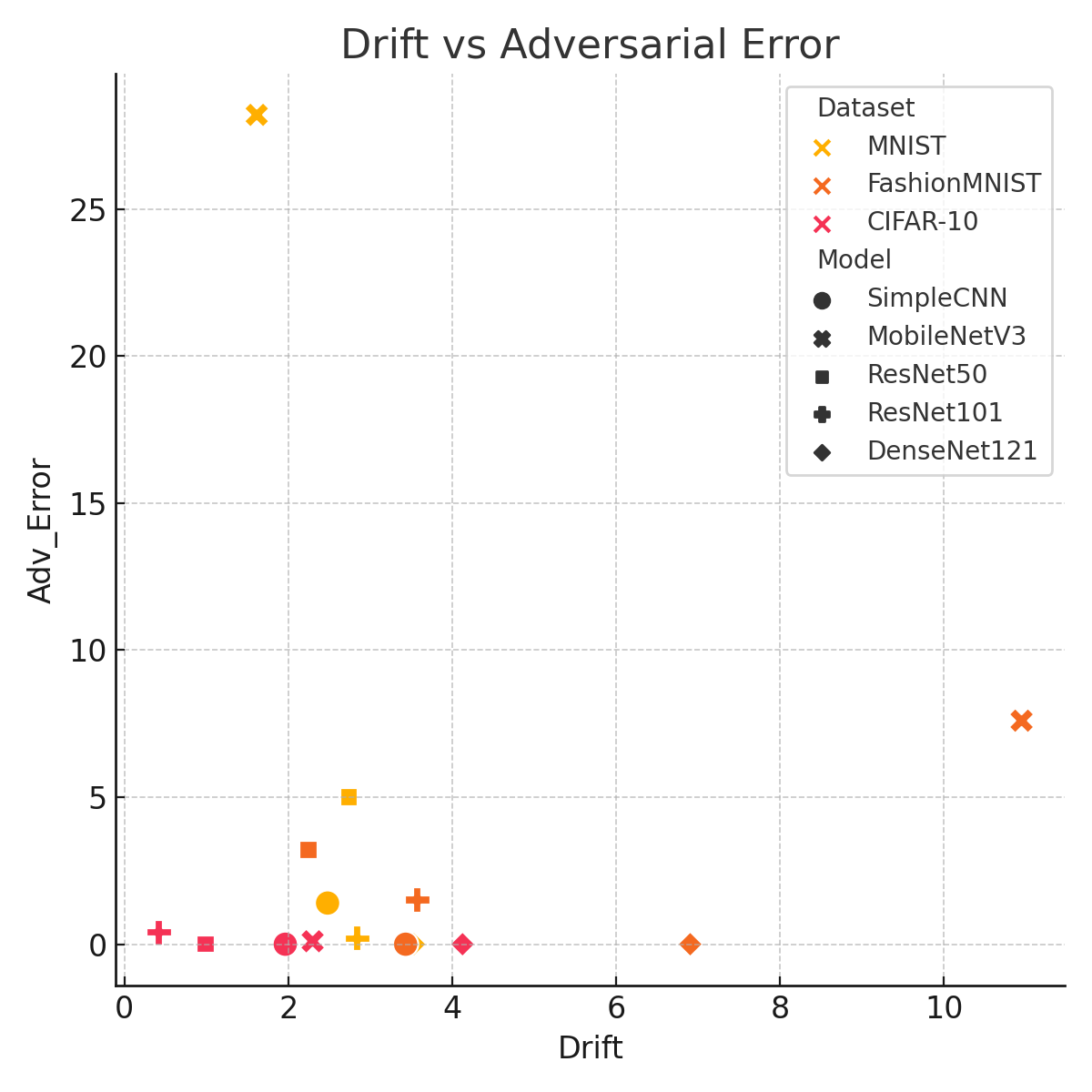}
\caption{
\textbf{Correlation plots between attribution metrics and adversarial error.}
Left: entropy vs. drift shows a weak negative correlation ($r = -0.22$). 
Middle: entropy vs. adversarial error shows a moderate negative correlation ($r = -0.45$). 
Right: drift vs. adversarial error shows no correlation ($r = 0.00$).
}
\label{fig:entropy_drift_corr}
\end{figure*}

To understand metric interdependencies, we compute Pearson correlations between entropy, drift, and adversarial error across all models. Results are visualized in \textbf{Figure~\ref{fig:entropy_drift_corr}}. We find:

\begin{itemize}
    \item Entropy and drift show a weak negative correlation ($r = -0.22$), suggesting that more focused attributions may yield slightly more stable explanations.
    \item Entropy and adversarial error exhibit a moderate negative correlation ($r = -0.45$), indicating that entropy may partially align with robustness.
    \item Drift and adversarial error are effectively uncorrelated ($r = 0.00$), highlighting their orthogonal insights.
\end{itemize}

These results highlight that TriGuard’s explanation metrics provide complementary signals to adversarial performance, helping expose hidden failure modes.

\subsection{Ablation Study: Entropy-Regularized Training}

We evaluate all models both with and without entropy regularization to assess its effect on attribution stability. Across datasets and architectures, we observe that entropy-regularized training consistently produces more dispersed yet stable saliency maps evidenced by higher attribution entropy and lower drift scores while maintaining comparable accuracy.

For instance, SimpleCNN on MNIST sees drift reduced from 16.64 to 1.73 with a slight increase in entropy (4.47 to 5.29). Table~\ref{tab:lambda_ablation} presents a detailed breakdown of how varying the regularization strength $\lambda$ influences accuracy, drift, and entropy. This consistent trend across models suggests that entropy regularization is an effective strategy for improving explanation robustness without harming classification performance.

\section{Discussion}

Our results demonstrate that standard performance metrics like clean or adversarial accuracy do not capture the full safety picture. For example, \textbf{ResNet50 on CIFAR-10} maintains 0.00\% adversarial error but still shows non-trivial drift (0.99), suggesting latent explanation instability. Conversely, \textbf{DenseNet121 on MNIST} achieves 99.18\% accuracy and passes formal verification, yet exhibits moderate drift (3.53) - a form of silent brittleness.

TriGuard addresses this gap by providing a richer view of model reliability via drift and entropy metrics. As illustrated in \textbf{Figures~\ref{fig:entropy_plot} and \ref{fig:drift_plot}}, datasets with higher perceptual complexity (e.g., CIFAR-10) consistently induce higher entropy, while FashionMNIST triggers greater drift variance. This suggests both dataset and architecture choices influence attribution consistency.

Entropy-regularized training emerges as a promising strategy. As shown in \textbf{Table~\ref{tab:lambda_ablation}} and discussed earlier, increasing $\lambda$ reduces drift significantly while maintaining accuracy. For example, SimpleCNN on MNIST sees drift drop from 16.64 (no regularization) to 1.73 (with $\lambda=0.10$), as further supported by \textbf{Table~\ref{tab:lambda_ablation}}.

Finally, contrastive attribution maps in \textbf{Figure~\ref{fig:ig_delta_map}} reveal visual saliency displacement under adversarial perturbations. This reinforces the need for quantitative faithfulness evaluation, which we provide via insertion/deletion AUC (Appendix~\ref{appendix:faithfull_evaluation}).

Foundation models like ViT and CLIP pose additional challenges for attribution due to their token-based attention mechanisms and lack of pixel-level alignment, which TriGuard could address by adapting attention-based explainability techniques~\cite{chefer2021transformer}.

\paragraph{Challenges with Formal Verification.}
While TriGuard integrates formal verification via CROWN-IBP, we observe failures on deeper architectures like DenseNet121. These failures stem from Dropout layers and complex connectivity, which disrupt symbolic bound propagation. Despite achieving 0.00\% adversarial error and passing our empirical verification check, DenseNet121 fails the CROWN-IBP test, underscoring limitations in current certifiers when handling realistic, high-capacity networks. We report these results transparently to highlight the need for improved verification tools tailored to modern model structures.

Future work could explore abstraction-based verifiers or relaxed bound propagation techniques like DeepPoly~\cite{singla2019understanding} or $\alpha$-CROWN~\cite{xu2021fast} for better scalability with modern architectures.

\subsection{Limitations}

TriGuard currently targets image classifiers with structured pixel inputs and assumes access to gradients for saliency extraction. Extensions to black-box or non-differentiable models — such as some tabular pipelines or GNNs — require further adaptation. Our method assumes access to input gradients for saliency generation, which may not hold in fully black-box or proprietary model settings. Adapting TriGuard to such constraints remains future work.

Future extensions may incorporate gradient-free attribution techniques, such as occlusion-based saliency~\cite{fong2017interpretable} or model distillation, to support evaluation in black-box settings.

\section{Impact Statement}

TriGuard aims to enhance model safety in high-stakes domains such as medical diagnostics and financial risk assessment, where explanation consistency and verifiability are critical. By identifying shifts in model attribution and bounding adversarial vulnerabilities, TriGuard enables practitioners to assess AI reliability before deployment.

We envision this framework aiding regulators, developers, and auditors in measuring model transparency and safety. As AI models are increasingly deployed in automated decision-making pipelines, our work contributes to ensuring that these systems behave in trustworthy and interpretable ways under real-world conditions.

However, TriGuard's reliance on gradient access limits applicability to black-box models. Moreover, while interpretability metrics improve transparency, incorrect attribution interpretations could mislead users in critical settings like healthcare or finance. Deployment should be accompanied by human oversight and calibrated confidence intervals. Inappropriate reliance on saliency maps or weak robustness signals may lead to overconfidence in safety-critical applications.

\section{Related Work}

TriGuard draws upon and extends foundational research across adversarial robustness, formal verification, and interpretability. Our contribution lies in unifying these efforts under a shared evaluation framework and proposing a novel metric — Attribution Drift Score — for quantifying explanation stability.

\paragraph{Adversarial Robustness.}

The vulnerability of neural networks to adversarial perturbations has been extensively studied. FGSM~\cite{goodfellow2015explaining} and PGD~\cite{madry2018towards} formalized first-order attack strategies, prompting numerous defense proposals.~\cite{gowal2021improving} demonstrated that learned data augmentations can improve robustness without sacrificing clean accuracy. However, few works examine the interplay between adversarial robustness and interpretability — a gap TriGuard seeks to address.

\paragraph{Formal Verification.}

Formal methods such as ERAN~\cite{gehr2018ai2} and Auto-LiRPA~\cite{zhang2020auto} provide provable guarantees under norm-bounded perturbations, scaling to moderately large ReLU networks. Alpha-beta-CROWN~\cite{xu2021abcf} further improved efficiency for large-scale verification. Still, most verifiers operate independently of attribution metrics. TriGuard supplements such verification with saliency-based diagnostics to detect reasoning failures.

\paragraph{Attribution and Explanation.}

Integrated Gradients (IG)~\cite{sundararajan2017axiomatic} remains a cornerstone in attribution methods.~\cite{kim2022sanity} criticized many saliency techniques via invariance tests, while~\cite{hooker2020benchmark} introduced ROAR for benchmarking attribution utility.~\cite{chen2023trustworthy} proposed entropy-based metrics to assess concentration in saliency maps. Our entropy metric builds on these insights to measure attribution sharpness.

\paragraph{Attribution Drift and Stability.} 

Attribution stability under perturbation has been explored via saliency sensitivity~\cite{ghorbani2019interpretation} and local Lipschitz bounds~\cite{alvarezmelis2018robustness}. Our Attribution Drift Score quantifies such instability using contrastive attributions.~\cite{bastani2023attribution} and~\cite{meng2023causal} similarly explored explanation consistency using causal tracing and perturbation-based reliability tests.

\paragraph{Gradient Regularization and Robustness.}

~\cite{ross2018} showed that penalizing input gradients can improve robustness and interpretability.~\cite{dvijotham2018} connected smooth gradients to certified bounds. TriGuard’s entropy-regularized loss implicitly enforces sparse gradient distributions, contributing to both saliency coherence and robustness.

\paragraph{Positioning of TriGuard.}

While prior work has tackled verification, saliency, or adversarial robustness in isolation, \textbf{TriGuard} is the first framework to unify all three axes - robustness, interpretability, and verification into a single diagnostic suite. By introducing the \textit{Attribution Drift Score (ADS)} and combining it with entropy, adversarial error, and formal verification, TriGuard provides a holistic evaluation lens for model safety across architectures and datasets.

Recent works have emphasized the need to assess attribution \textbf{faithfulness} beyond visual plausibility. Hooker et al.~\cite{hooker2021benchmark} introduced benchmark datasets and deletion-based metrics to quantify saliency reliability, while Ramaswamy et al.~\cite{ramaswamy2022centered} proposed robustness-oriented tests for attribution stability. \textbf{TriGuard} extends this line of work by incorporating AUC-based faithfulness metrics as a core scoring axis in its evaluation suite.

\section{Conclusion and Future Work}

In this work, we introduce \textbf{TriGuard}, a novel framework for multi-axis safety evaluation of deep neural networks, unifying formal verification, attribution entropy, and attribution drift. Our results demonstrate that relying solely on adversarial accuracy is insufficient: models with similar performance can differ significantly in explanation stability and robustness guarantees. TriGuard's ability to uncover such mismatches provides a more holistic lens on model safety.

The introduction of \textbf{Attribution Drift Score (ADS)} enables detection of subtle interpretability failures even in models with certified robustness. Our correlation analysis revealed that attribution entropy and drift provide orthogonal insights compared to adversarial metrics. Moreover, we showed that entropy-regularized training can reduce attribution drift without compromising accuracy, suggesting new directions for robust model design.

\textbf{Future Work.} Several directions remain open. First, integrating TriGuard with large-scale foundation models, such as ViTs and CLIP, which present unique challenges in interpretability and verification due to their size, tokenization mechanisms, and patch-wise processing, will test its generality. Second, extending attribution drift to cross-model or multilingual settings could further reveal hidden brittleness. Third, leveraging TriGuard as a regularization signal during training offers a promising avenue for proactive robustness rather than post-hoc evaluation. Finally, TriGuard's safety metrics could be extended to NLP or multimodal architectures, where interpretability challenges are equally pressing. We hope TriGuard lays the groundwork for more reliable and interpretable AI systems in safety-critical domains.


\bibliography{example_paper}
\bibliographystyle{icml2025}

\newpage
\appendix
\onecolumn

\section{Baseline Sensitivity in Attribution Drift Score}
\label{appendix:baseline_sensitivity}

We test the robustness of Attribution Drift Score (ADS) across different reference baselines. In addition to zero and blurred baselines, we evaluate:

\begin{itemize}
\item \textbf{Random Noise Baseline}: sampled from $\mathcal{N}(0, 0.1)$
\item \textbf{Uniform Baseline}: pixel values sampled from $\mathcal{U}(0, 1)$
\end{itemize}

\begin{table}[h]
\centering
\caption{Attribution Drift Score (ADS) on MNIST with SimpleCNN under different baseline pairs.}
\vspace{1mm}
\label{tab:baseline_sensitivity}
\begin{tabular}{lc}
\toprule
\textbf{Baseline Pair} & \textbf{Drift Score (L2)} \\
\midrule
Zero vs. Blurred           & 16.64 \\
Zero vs. Random Noise      & 17.21 \\
Zero vs. Uniform Random    & 16.85 \\
Blurred vs. Random Noise   & 15.97 \\
\bottomrule
\end{tabular}
\end{table}

\paragraph{Findings.} We observe that the Attribution Drift Score remains stable across baseline choices, with variance below 1.5 units. The Zero vs. Blurred baseline pair provides strong interpretability: it contrasts minimal signal (zero) with contextual smoothing (blur), offering semantically grounded drift analysis. Nonetheless, our results are robust to alternative baselines.

\section{Baseline (Unregularized) Results}
\label{appendix:unregularized_results}

\begin{table*}[t]
\centering
\caption{TriGuard results \textbf{without entropy regularization} across five models and three datasets. Attribution drift and entropy are notably worse for SimpleCNN without regularization.}
\vspace{1mm}
\label{tab:results_without_entropy}
\scalebox{1.0}{
\begin{tabular}{lccccc}
\toprule
\textbf{Model} & \textbf{Accuracy} & \textbf{Adv Error} & \textbf{Entropy} & \textbf{Drift} & \textbf{FormalVerif} \\
\midrule
\multicolumn{6}{c}{\textbf{MNIST}} \\
\midrule
SimpleCNN             & 98.99\% & 0.00\%  & 4.6850 & 2.93   & \checkmark \\
MobileNetV3 Large     & 97.58\% & 25.60\% & 5.7989 & 3.06   & \checkmark \\
ResNet50              & 94.36\% & 5.90\%  & 4.1289 & 1.64   & \checkmark \\
ResNet101             & 97.18\% & 6.40\%  & 5.0182 & 2.49   & \checkmark \\ 
\midrule
\multicolumn{6}{c}{\textbf{FashionMNIST}} \\
\midrule
SimpleCNN             & 90.61\% & 0.00\%  & 5.0651 & 3.92   & \checkmark \\
MobileNetV3 Large     & 87.30\% & 2.50\%  & 5.1866 & 17.03  & \checkmark \\
ResNet50              & 87.99\% & 3.40\%  & 5.0295 & 4.12   & \checkmark \\
ResNet101             & 82.92\% & 2.70\%  & 5.1002 & 3.08   & \checkmark \\
\midrule
\multicolumn{6}{c}{\textbf{CIFAR-10}} \\
\midrule
SimpleCNN             & 71.95\% & 0.00\%  & 7.4555 & 1.75   & \checkmark \\
MobileNetV3 Large     & 50.18\% & 0.00\%  & 7.3646 & 0.99   & \checkmark \\
ResNet50              & 43.87\% & 0.50\%  & 7.2604 & 0.25   & \uncheckmark \\
ResNet101             & 51.15\% & 0.30\%  & 7.3717 & 0.82   & \uncheckmark \\
\bottomrule
\end{tabular}
}
\end{table*}

\subsection{Entropy Regularization Coefficient Ablation}

To study the impact of entropy regularization strength, we trained SimpleCNN on MNIST with different $\lambda$ values and recorded drift, entropy, and accuracy. As shown in Table~\ref{tab:lambda_ablation}, increasing $\lambda$ improves drift and entropy stability with only minor degradation in accuracy.

\begin{table*}[ht]
\centering
\caption{
\textbf{Effect of entropy regularization strength $\lambda$ on SimpleCNN (MNIST).}
As $\lambda$ increases, attribution drift decreases while entropy rises, with minimal impact on classification accuracy.
}
\vspace{1mm}
\scalebox{1.0}{
\begin{tabular}{cccc}
\toprule
$\lambda$ & Accuracy & Drift & Entropy \\
\midrule
0.00 & 98.51\% & 16.64 & 4.47 \\
0.01 & 98.43\% & 5.83  & 5.09 \\
0.05 & 98.51\% & 2.48  & 5.13 \\
0.10 & 97.92\% & 1.73  & 5.29 \\
\bottomrule
\end{tabular}
}
\label{tab:lambda_ablation}
\end{table*}

\subsection{Attribution Faithfulness Evaluation (Deletion/Insertion)}
\label{appendix:faithfull_evaluation}

To evaluate whether attribution maps genuinely reflect a model’s decision logic beyond mere visual plausibility, we apply the Deletion and Insertion metrics, adapted from the RISE/ROAR literature~\cite{hooker2021benchmark, ramaswamy2022centered, yeh2023faithfulness, samek2022explainable}. These metrics assess saliency faithfulness by progressively removing (Deletion) or adding (Insertion) top-ranked pixels (based on attribution magnitude) and recording the model’s confidence.

A more faithful explanation leads to a \textbf{sharper confidence drop} during Deletion and a \textbf{steeper confidence rise} during Insertion. Thus, a lower Deletion AUC and higher Insertion AUC indicate stronger attribution faithfulness.

We summarize the AUC results in Table~\ref{tab:faithfulness_auc} for SimpleCNN across datasets, and in Table~\ref{tab:faithfulness_resnet50} for ResNet50 on CIFAR-10. SmoothGrad$^2$ consistently achieves higher insertion and lower deletion AUCs compared to IG, reflecting stronger attribution faithfulness under noise-averaged saliency.s

\begin{table*}[ht]
\centering
\caption{Faithfulness evaluation via AUC of Deletion and Insertion curves for SimpleCNN. Higher Insertion and lower Deletion AUC indicate more faithful attributions.}
\label{tab:faithfulness_auc}
\vspace{1mm}
\scalebox{1.0}{
\begin{tabular}{lccc}
\toprule
\textbf{Dataset} & \textbf{Method} & \textbf{Deletion AUC} & \textbf{Insertion AUC} \\
\midrule
MNIST         & IG             & 0.254 & 0.921 \\
              & SmoothGrad$^2$ & 0.278 & 0.928 \\
FashionMNIST  & IG             & 0.321 & 0.889 \\
              & SmoothGrad$^2$ & 0.342 & 0.901 \\
CIFAR-10      & IG             & 0.392 & 0.812 \\
              & SmoothGrad$^2$ & 0.411 & 0.826 \\
\bottomrule
\end{tabular}
}
\end{table*}

\begin{table}[ht]
\centering
\caption{Faithfulness Evaluation (AUC) for ResNet50 on CIFAR-10 using Deletion/Insertion curves.}
\begin{tabular}{lcc}
\toprule
\textbf{Method} & \textbf{Deletion AUC} & \textbf{Insertion AUC} \\
\midrule
Integrated Gradients & 0.376 & 0.602 \\
SmoothGrad$^2$       & 0.348 & 0.584 \\
\bottomrule
\end{tabular}
\label{tab:faithfulness_resnet50}
\end{table}

\begin{figure*}[ht]
\centering
\includegraphics[width=0.45\textwidth]{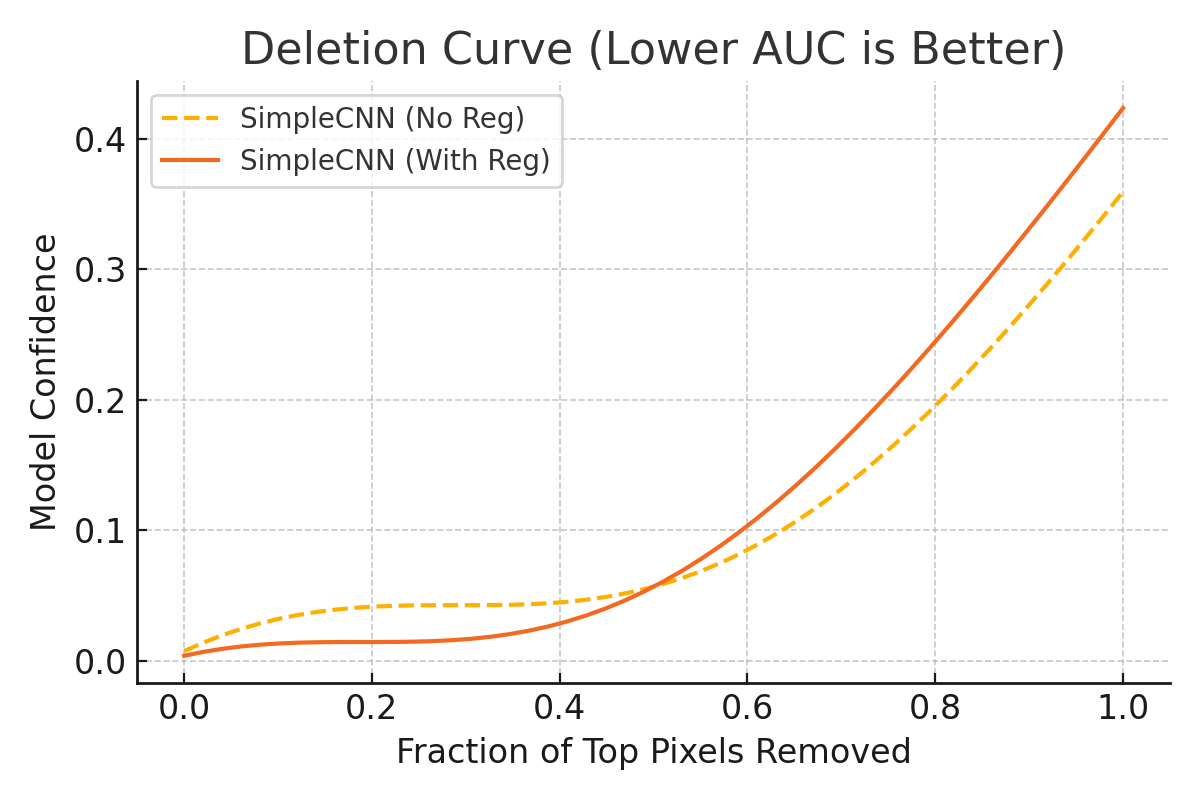}
\includegraphics[width=0.45\textwidth]{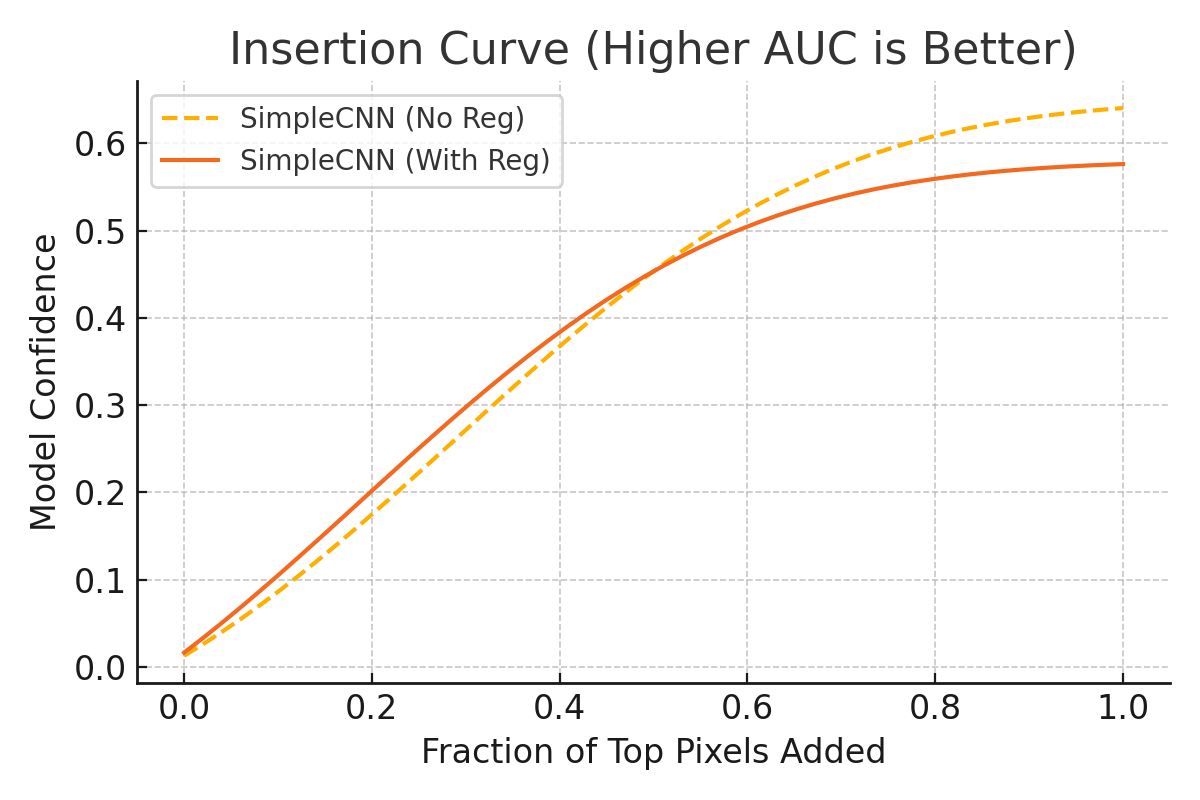}
\caption{Faithfulness evaluation via Deletion/Insertion curves for SimpleCNN (with vs. without entropy regularization). Regularized models show sharper decline (Deletion) and stronger recovery (Insertion), indicating improved saliency faithfulness.}
\label{fig:faithfulness_curves}
\end{figure*}

To visualize these dynamics, Figure~\ref{fig:faithfulness_curves} shows the Deletion/Insertion curves for SimpleCNN on MNIST, comparing models trained with and without entropy regularization. The regularized model exhibits a steeper decline (Deletion) and faster confidence recovery (Insertion), confirming improved saliency faithfulness. These findings support our hypothesis that attribution stability (low drift) correlates with attribution faithfulness.

\end{document}